\documentclass[conference]{IEEEtran}
\usepackage{amsfonts}
\usepackage{times}
\usepackage{graphicx}
\usepackage{latexsym}
\usepackage{dsfont}
\usepackage{amssymb}
\usepackage{amsmath}
\usepackage{cite}
\usepackage{verbatim}
\usepackage{subfigure}

\newcommand{\figref}[1]{{Fig.}~\ref{#1}}

\def\bb0{{\mathbb{0}}}

\def\ba{{\mathbf{a}}}
\def\bb{{\mathbf{b}}}

\def\b0{{\mathbf{0}}}

\def\bX{{\mathbf{X}}}

\def\sf0{{\mathsf{0}}}

\usepackage{epstopdf}
\usepackage{enumerate}
\usepackage{algorithmicx}
\usepackage{algorithm}
\usepackage{amsmath}
\usepackage[noend]{algpseudocode}
\usepackage{float}
\usepackage{hyperref}
\usepackage{color}
\usepackage{makeidx}
\usepackage{bbm}
\usepackage{graphicx}
\usepackage{lipsum}
\usepackage{subfigure}
\usepackage{tablefootnote}
\usepackage{multirow}
\usepackage{multicol}
\usepackage{balance}
\usepackage{booktabs}

\def\j{\mathrm{j}}

\newcommand{\comm}[1]{}

\begin{document}

\title{Vision-Aided Dynamic Blockage Prediction  for \\ 6G Wireless Communication Networks}
\author{Gouranga Charan, Muhammad Alrabeiah, and Ahmed Alkhateeb\\ Arizona State University, Emails: \{gcharan, malrabei, alkhateeb\}@asu.edu}

\maketitle

\begin{abstract}
Unlocking the full potential of millimeter-wave and sub-terahertz wireless communication networks hinges on realizing unprecedented low-latency and high-reliability requirements. The challenge in meeting those requirements lies partly in the sensitivity of signals in the millimeter-wave and sub-terahertz frequency ranges to blockages. One promising way to tackle that challenge is to help a wireless network develop a sense of its surrounding using machine learning. This paper attempts to do that by utilizing deep learning and computer vision. It proposes a novel solution that proactively predicts \textit{dynamic} link blockages. More specifically, it develops a deep neural network architecture that learns from observed sequences of RGB images and beamforming vectors how to predict possible future link blockages. The proposed architecture is evaluated on a publicly available dataset that represents a synthetic dynamic communication scenario with multiple moving users and blockages. It scores a link-blockage prediction accuracy in the neighborhood of 86\%, a performance that is unlikely to be matched without utilizing visual data. 
\end{abstract}


\section{Introduction} \label{sec:Intro}

Millimeter-wave (mmWave) and sub-terahertz communications are becoming the dominant directions for modern and future wireless networks \cite{HeathJr2016, Rappaport2019}. With their large bandwidth, they have the ability to satisfy  the high data rate demands of several applications such as wireless Virtual/Augmented Reality (VR/AR) and autonomous driving. Communication in these bands, however, faces several challenges at both the physical and network layers. One of the key challenges stems from the sensitivity of mmWave and terahertz signal propagation to blockages. This requires these systems to rely heavily on maintaining line-of-sight (LOS) connections between the base stations and users.  The possibility of blocking these LOS links by stationary or dynamic blockages can highly affect the reliability and latency of mmWave/terahertz systems.

The key for overcoming the link blockage challenges lies in enabling the wireless system to develop a sense of its surrounding. This could be accomplished using machine learning. Many recently published studies have shown that using wireless sensory data (e.g., channels, received power, etc.), a machine learning model can efficiently differentiate LOS and Non-LOS (NLOS) links \cite{LSTM_blk}, \cite{Sub6PredMmWave}, \cite{BlockagePred}. The former two studies address the link blockage problem form a \textit{reactive} perspective; the sensory data is first acquired, and, then, the status of the current link is predicted. Despite its appeal, the reactive link-blockage prediction does not satisfy the low-latency and high-reliability requirements of future wireless networks. The work in \cite{BlockagePred} takes a step forward towards a \textit{proactive} treatment of the problem. It studies proactive blockage prediction for a single-moving mmWave user in the presence of \textit{stationary} blockages. Again, despite its appeal, it still falls short in meeting the latency and reliability requirements.

In this paper and inspired by the recently proposed Vision-Aided Wireless Communication (VAWC) framework in \cite{CamPredBeam} and \cite{ViWi}, the proactive link-blockage problem in fully dynamic environments is addressed using deep learning and a fusion of visual and wireless data. Images and video sequences usually speak volumes about the environment they are depicting. The empirical evidence in \cite{CamPredBeam} supports that notion; a deep learning model is able to perform simple mmWave beam and blockage prediction tasks using visual data. As such, a fusion of visual and wireless data is viewed as the right type of sensory data that could facilitate the learning of the more complex task of proactive link-blockage prediction. 
Based on this motivation, we develop a novel deep learning architecture to utilize sequences of observed RGB images (frames) and beamforming vectors and learn proactive link-blockage prediction. The proposed architecture harnesses the power of Convolutional Neural networks (CNNs) and Recurrent Neural Networks (RNNs). The former excels in performing visual tasks such as image classification \cite{resnet} and object detection \cite{Yolo} while the later defines the stat-of-the-art in sequence modeling, as evident by the performance of RNNs in speech recognition \cite{SpeechRecog, SpeechRecog2}. Experimental results show that the proposed architecture is capable of learning the prediction task effectively, achieving a noticeable gain over the models that rely solely on wireless data.

The rest of this paper is organized as follows. Section \ref{sec:sys_ch_mod} introduces the system and channel models. Section \ref{sec:prob_form} provides a formal description of the proactive link-blockage prediction problem. Following the problem definition, Section \ref{sec:prop_sol} presents a detailed description of the proposed architecture, while Section \ref{sec:exp_set}  details  the experimental setup used to evaluate the performance of the architecture. The performance evaluation, then, is presented in Section \ref{sec:perf_eval}. Finally, Section \ref{sec:conc} wraps up the paper with some concluding remarks.

\section{System and Channel Models} \label{sec:sys_ch_mod}
To illustrate the potential of deep learning and VAWC in mitigating the link blockage problem, this work considers a high-frequency communication system where the basestation utilizes an RGB camera to monitor its environment. The following two subsections provide a brief description of that system and its adopted channel model. 

\subsection{System model}
\begin{figure}
    \centering
    \includegraphics[width=\linewidth]{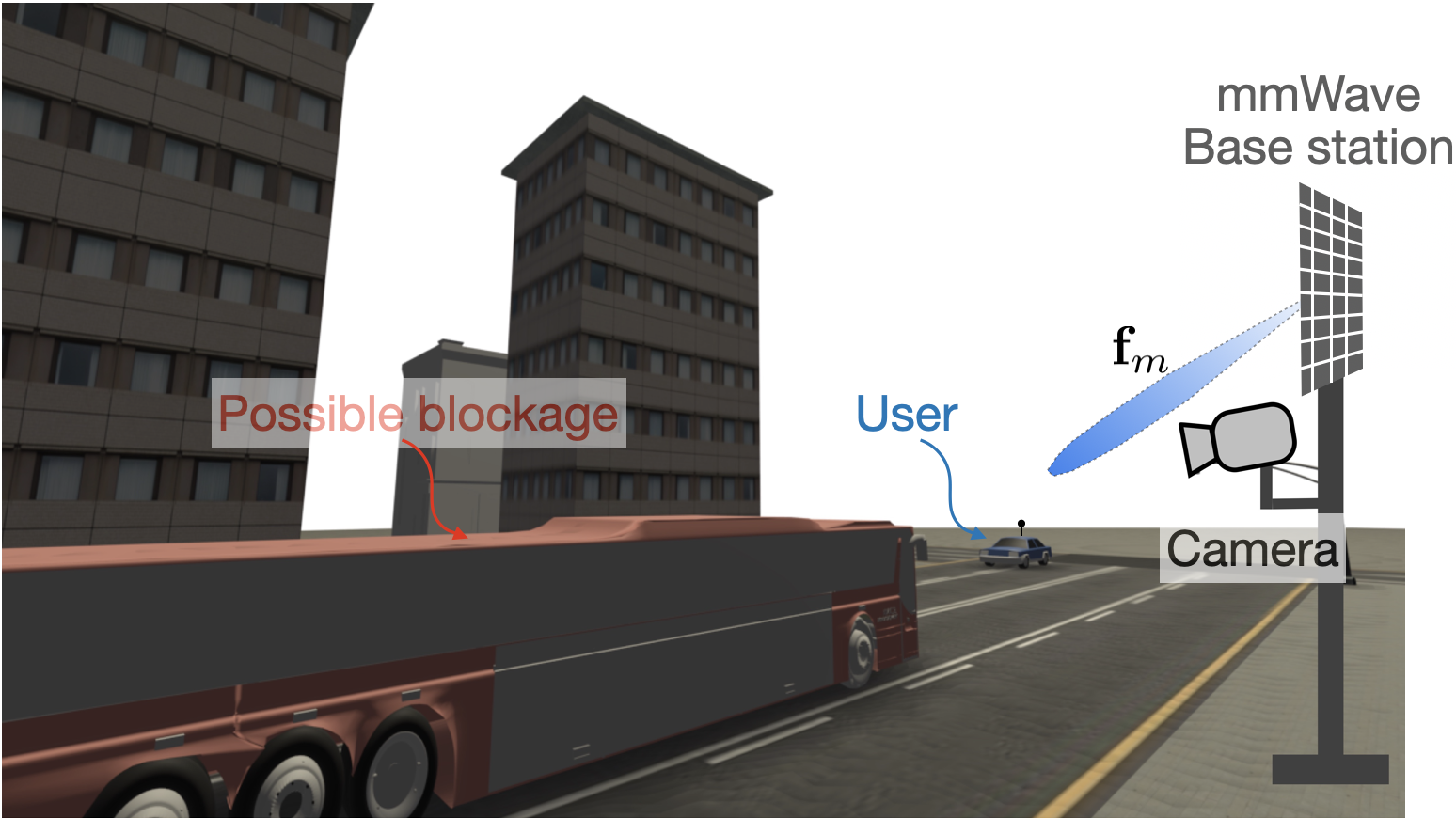}
    \caption{A mmWave basestation equipped with a mmWave array and a camera serving a mobile user. The user is moving alnog the street and heading towards a possible blockage (bus).}
    \label{fig:sys_mod}
\end{figure}

A depiction of the system considered in this paper is shown in  \figref{fig:sys_mod}. It has a mmWave basestation equipped with $M$ antennas and an RGB camera. The communication system adopts OFDM  and uses a predefined beamforming codebook $\boldsymbol{\mathcal F}=\{\mathbf f_m\}_{m=1}^{Q}$, where $\mathbf{f}_m \in \mathbb C^{M\times 1}$ and $Q$ is the total number of beamforming vectors in the codebook. For any mmWave user in the wireless environment, its received downlink signal is given by:
\begin{equation}
    y_{u,k} = \mathbf h_{u,k}^T \mathbf f_m x + n_k,
\end{equation}
where $y_{u,k}\in \mathbb C$ is the received signal of the $u$th user at the $k$th subcarrier, $\mathbf h_{u,k} \in \mathbb C^{M\times 1}$ is the  channel between the BS and the $u$th user at the $k$th subcarrier, $x\in \mathbb C$ is a transmitted complex symbol, and $n_k$ is a noise sample drawn from a complex Gaussian distribution $\mathcal N_\mathbb C(0,\sigma^2)$.

\subsection{Channel model}
A geometric channel model is assumed in this work, in which the channel vector of the $u$th user at the $k$th subcarrier is given by:
 \begin{equation}
\mathbf{h}_{u,k} = \sum_{d=0}^{D-1} \sum_{\ell=1}^L \alpha_\ell e^{- \j \frac{2 \pi k}{K} d} p\left(dT_\mathrm{S} - \tau_\ell\right) \ba\left(\theta_\ell, \phi_\ell\right),
\end{equation} 
where $L$ is number of channel paths, $\alpha_\ell, \tau_\ell, \theta_\ell, \phi_\ell$ are the path gains (including the path-loss), the delay, the azimuth angle of arrival, and the elevation angle of arrival, respectively, of the $\ell$th channel path. $T_\mathrm{S}$ represents the sampling time while $D$ denotes the cyclic prefix length (assuming that the maximum delay is less than $D T_\mathrm{S}$).

\section{Problem formulation} \label{sec:prob_form}

The primary objective of this paper is to utilize sequences of RGB images and beam indices and develop a machine learning model that learns to predict link blockages proactively, i.e., transitions from LOS to NLOS. Formally, this learning problem could be posed as follows. For any user $u$ in the environment, a sequence of image and beam-index\footnote{Since the system model assumes a predefined beamforming codebook, the indecies of those beams are used instead of the complex-valued vectors themselves.} pairs is observed over a time interval of $r$ instances. At any time instance $t\in \mathbb Z$, that sequence is given by:
\begin{equation}
    {\mathcal S}_{u}[t] = \{ (\bX_u[i], b_u[i]) \}_{i = t-r+1}^{t},
\end{equation}
where $b_{u}[i]$ is the index of the beamforming vector in codebook $\boldsymbol{\mathcal{F}}$ used to serve user $u$ at the $i$th time instance, $\bX[i] \in \mathbb{R}^{W \times H \times C}$ is an RGB image of the environment taken at the $i$th time instance, $W$, $H$, and $C$ are respectively the width, height and the number of color channels for the image, and $r\in \mathbb Z$ is the extent of the observation interval. Furthermore, let $s_u[t+1] \in \{0,1\}$ represent the user's link status at instance $t+1$ where $0$ and $1$ respectively represent the LOS and NLOS link statuses. The objective, then, is to utilize the sequence $\mathcal S_u[t]$ and predict the link status $s_u[t+1]$ with high fidelity, i.e., if $\hat s_u[t+1]$ is the estimated status, then this estimation must maintain a high success probability $\mathbb P(\hat s_u[t+1] = s_u[t+1]|\mathcal S_u [t])$.

That objective is attained using a machine learning model. It is developed to learn a prediction function $f_{\Theta}(\mathcal S_u[t+1])$ that takes in the observed image-beam pairs and produces $\hat s_u[t+1]$. This function is parameterized by a set $\Theta$ representing the model parameters and learned from a dataset of labeled sequences, i.e., $\mathcal D = \{(\mathcal S_u[t], s_u[t+1])\}_{u=1}^U$where each pair has the observed sequence and its \textit{groundtruth} future link status\footnote{It is importance to notice here that all sequences $\mathcal S_u[t]$ are indexed by $t$ for simplicity of exposition and not to say that they are all captured at the same time instance. In fact, sequences are captured at different time instances and only share the same observation interval $r$.}. Following the formulation convention in machine learning \cite{PatternRecog,DLBook} and to maintain the high fidelity requirement, the prediction function is learned such that it maximizes the joint success probability of all data samples in $\mathcal D$. This is expressed by:
\begin{equation}\label{obj}
\underset{f_{\Theta}(\mathcal S_u[t])}{\text{max}} \quad  \prod_{u=1}^{U} \mathbb P(\hat s_u[t+1] = s_u[t+1]|\mathcal S_u [t]),
\end{equation}
where the joint probability in \eqref{obj} is factored out to convey the implicit assumption that for any user $u$, its success probability only depends on its observed sequence $\mathcal S_u[t]$.

 \begin{figure*}[t]
 	\centering
 	\includegraphics[width=.9\linewidth]{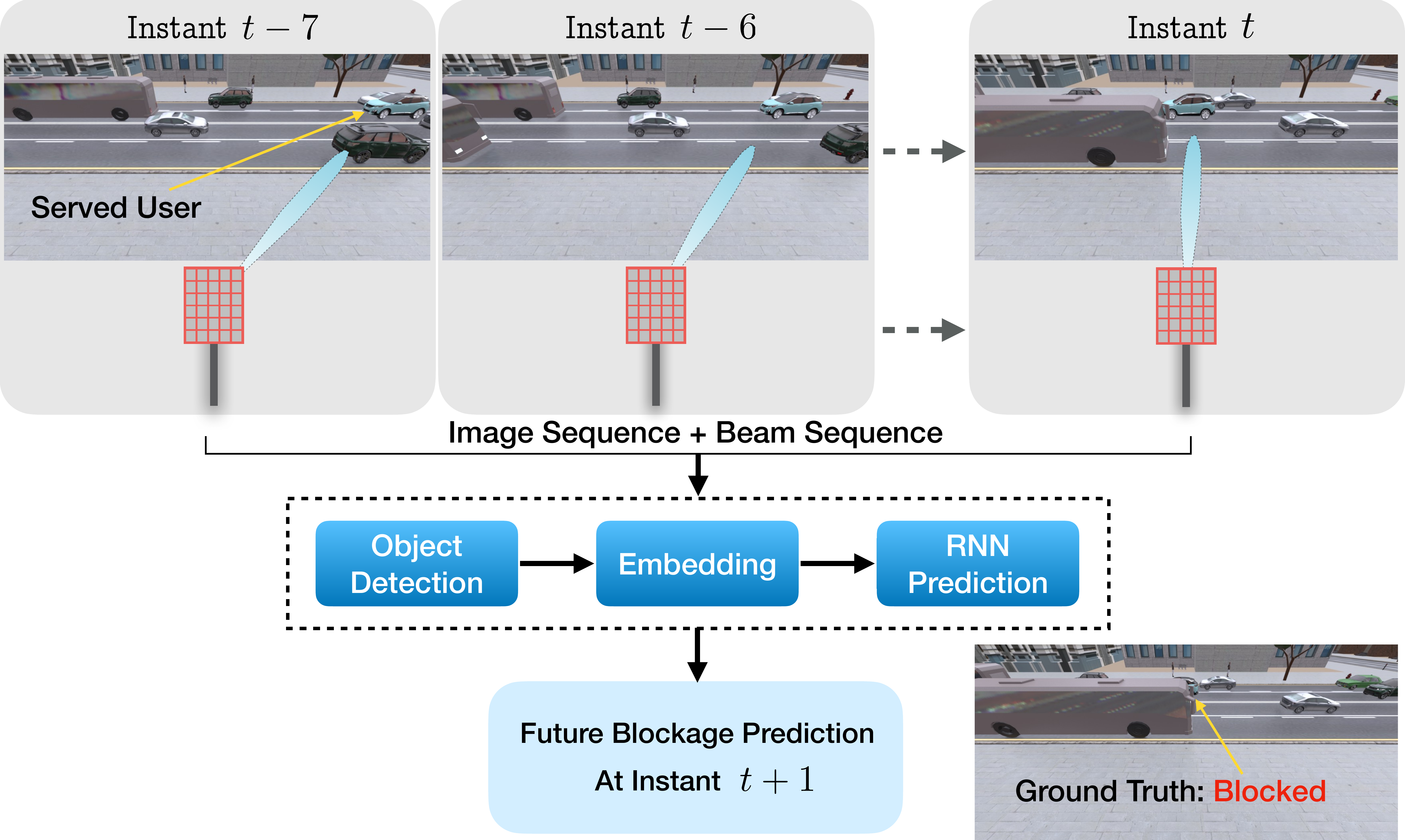}
 	\caption{An overall blockage diagram illustrating the problem and its solution. The deep learning model takes in a sequence of observed image-beam pairs and produces a prediction of the link status of the user in a future time instance.}
 \end{figure*}

\section{Proposed Solution} \label{sec:prop_sol}

Given the objective of predicting future link blockages, the prediction function could be viewed as a binary classifier separating the observed sequences $\mathcal S[t]$ into two classes, LOS/unblocked ($s[t+1] = 0$) and NLOS/blocked ($s[t+1]=1$) --note that the subscript $u$ is henceforth dropped for brevity. The function is envisioned to perform the following process. It first transforms the image-beam sequence into a sequence of high-dimensional features. Then, it learns the temporal relation among those features and encodes it into a single high-dimensional vector. The vector is, finally, fed to the classifier to predict the link status (class label). The proposed solution translates that abstract description into a three-component network, as illustrated in Fig.\ref{fig:CNN-RNN}. The following subsections provide a deeper dive into each component.

\subsection{Object Detector }
\label{subsec:obj_det}
\begin{figure*}[!htb]
	\centering
	\includegraphics[width=0.9\linewidth]{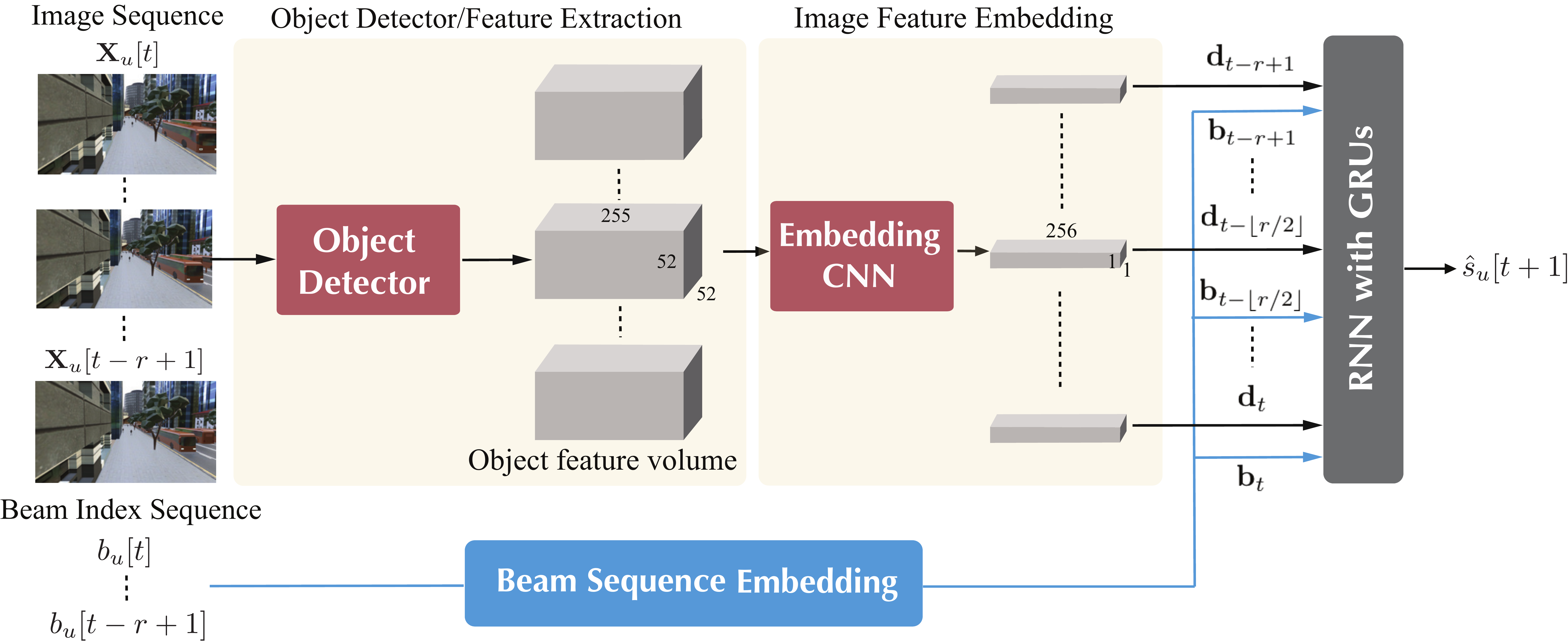}
	\caption{A block diagram showing the proposed neural network. It shows the three main components of the architecture: (i) the object detector, (ii) the embedding component with a CNN network and a beam-embedding block , and, finally, (iii) the recurrent prediction network. }
	\label{fig:CNN-RNN}
\end{figure*}

Recent advances in deep learning have shown CNNs to be the state-of-the-art networks for visual tasks like image classification \cite{resnet, vgg} and object detection \cite{Yolo, SSD} to name two. As such, the architecture of the proposed solution implements a CNN-based object detector as its first component. The job of this detector is to extract visual features pertaining to relevant objects in the environment like cars, people, buses, etc. Since object detection in itself is not the objective of this work, the detector is not equipped with an object classifier or a bounding-box predictor. Instead, it has a sequence of convolutional layers that extract a feature volume at the end. As shown in Fig.\ref{fig:CNN-RNN}, the detector is designed to sequentially process the images in the input sequence and extract their corresponding feature volumes.

The object detector in the proposed solution needs to meet two essential requirements: (i) detecting a wide range of objects, and (ii) producing quick and accurate predictions. These two requirements have been addressed in many of the state-of-the-art object detectors. A good example of a fast and accurate object-detection neural networks is the You Only Look Once (YOLO) detector, proposed first in\cite{Yolo} and then improved in \cite{Yolov2}. Therefore, instead of building and training a new object detector from scratch, the proposed solution utilizes a pre-trained YOLO network and integrates it into its architecture with some minor modifications; the detector is stripped off of its classifier and bounding-box predictor. Such choice allows the solution to extract rich feature volumes without the need to undergo extensive object-detection training, commonly associated with state-of-the-art detectors.

\subsection{Feature and Beam Embedding}
\label{subsec:feat_beam_emb}
The prediction function relies on dual-modality observed data, i.e., visual and wireless data. Although such data is expected to be rife with information, its dual nature brings about a heightened level of difficulty from the learning perspective. In an effort to overcome that difficulty, the proposed solution incorporates an embedding component that processes the extracted image feature volumes and the beam indices separately., as shown in the middle part of Fig.\ref{fig:CNN-RNN}. It transforms them to the same $N$-dimensional space before they are fed to the next component. For beam indices in the input sequence, the adopted embedding is simple and does not require any training. It generates a lookup table of $\left|\boldsymbol{\mathcal F}\right|$ real-valued vectors $\mathbf b_i\in \mathbb R^N$ where $i\in\{t-r+1, \dots, t\}$. The elements of each vector are randomly drawn from a Gaussian distribution with zeros mean and unity standard deviation.

For the extracted feature volumes, on the other hand, they are embedded using a trainable CNN network. It takes in one feature volume at a time and produces a feature vector $\mathbf d_i \in \mathbb R^N$ where $i\in\{t-r+1, \dots, t\}$, see the image feature embedding part in Fig.\ref{fig:CNN-RNN}. Since a volume embodies rich information about the objects in the environment, some of that information may not be relevant to the blockage prediction task. Hence, the CNN is designed to learn how to squeeze out the most relevant information in the volume and embed it in the $N$-dimensional vector space. The CNN implements three convolution layers, each of which has $255$ 3D filters with spatial dimensions of $5\times 5$ followed by a batch-normalization layer and max-pooling layer. The third convolution layer is followed by two fully-connected layers that produce the embedded feature.

\subsection{Recurrent prediction }
\label{subsec:rnn}
CNN networks inherently fail in capturing sequential dependencies in input data, thereby they are not expected to learn the relation among a sequence of embedded features. To overcome that, the third component of the proposed architecture utilizes Recurrent Neural Networks (RNNs) and perform future blockage prediction based on the learned relation among those features. In particular, the recurrent component has two layers of Gated Recurrent Units (GRU) separated by a dropout layer. These two layers are followed by a fully-connected layer that acts as a classifier. The recurrent component receives a sequence of length $2r$ of alternating image and beam embeddings, i.e., a sequence of the form $\{\mathbf d_{t-r+1}, \mathbf b_{t-r+1},\dots,\mathbf d_t, \mathbf b_t\}$, and hence, it implements $2r$ GRUs per layer. The output of the last unit in the second GRU layer is fed to the classifier to predict the future link status $\hat s[t+1]$.


\begin{table}[!t]
\caption{Design and Training Hyper-parameters}
\label{table}
\centering
\setlength{\tabcolsep}{5pt}
\renewcommand{\arraystretch}{1.2}
\begin{tabular}{|l|c|c|}
\hline
\multirow{9}{*}{Design}   & Embedding convolution kernel & $5\times 5$       \\ \cline{2-3}
                          & Number of kernels per layer & 256 \\ \cline{2-3}
						  & Kernel stride & 1 \\ \cline{2-3}
						  & Pooling kernel & $2\times 2$ \\ \cline{2-3}
						  & Pooling stride & 1 \\ \cline{2-3}
						  & Number of GRUs Per Layer ($2r$)   & $16$             \\  \cline{2-3}
                          & Embedding Dimension ($N$)        & $256$                        \\ \cline{2-3} 
                          & Hidden State Dimension         & $20$                         \\ \cline{2-3} 
                          & Number of classes  & $2$                        \\ \hline\hline 
\multirow{4}{*}{Training} & Optimizer                      & ADAM                       \\ \cline{2-3} 
                          & Learning Rate                  & $1 \times 10 ^{-3}$ \\ \cline{2-3} 
                          & Batch Size                     & $1000$                       \\ \cline{2-3} 
                          & Number of Training Epochs               & $50$                         \\ \hline
\end{tabular}
\label{tab_non_idealities}
\vspace{-4mm}
\end{table}


\section{Experimental Setup}\label{sec:exp_set}
The following few sections discuss the dataset used in our simulation, the evaluation metrics followed, and the neural network training. 
\subsection{Dataset}

\begin{figure}[t]
\centering
\includegraphics[width=.96\linewidth]{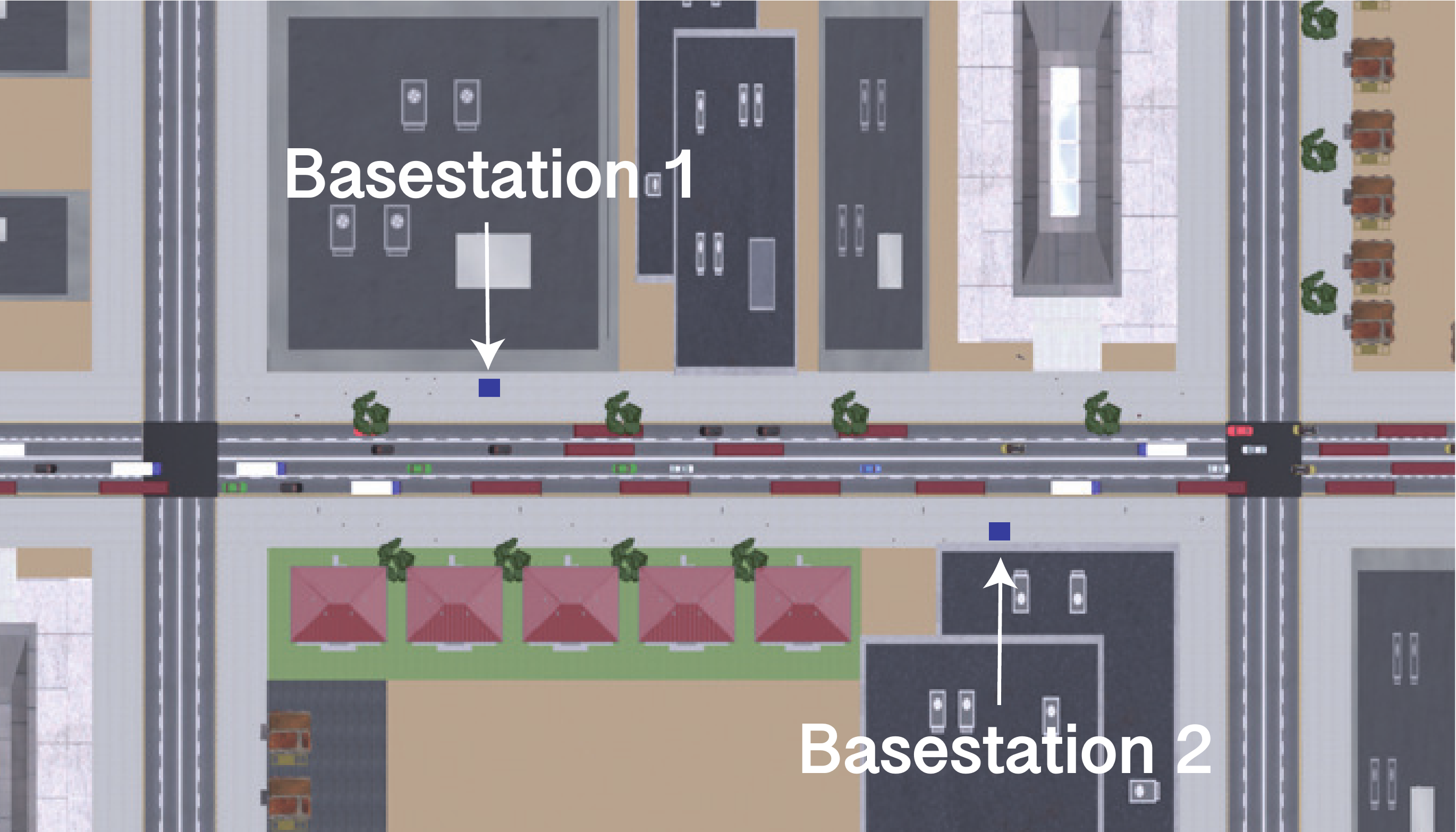}
\caption{A top-view of the simulation outdoor scenario. It is modeled after a busy downtown street with variety of moving and stationary objects, such as cars, buses, trucks, pedestrians, high-rises, trees, etc. The view also depicts the two basestations.}
\label{top_view}
\end{figure}

The extended ViWi-BT \cite{ViWi} dataset is used to evaluate the performance of the proposed solution. The dataset provides co-existing wireless and visual data. The dataset is an extension of the ViWi-BT challenge dataset presented in \cite{viwi_bt}. It comprises sequences of RGB frames, beam indices, and user link statuses. They are generated from a large simulation of a synthetic outdoor environment depicting a downtown street with multiple moving objects. Fig.\ref{top_view} shows a top view of that environment. It shows two small-cell mmWave basestations deployed opposite to each other and at different ends of the street. Each basestation is equipped with three differently-oriented cameras covering the whole street. More information about the dataset could be found at \cite{ViWi}.

The dataset has a little shy of $180$ thousand data samples (sequences). It is split into training and validation sets using a 70\%-30\% data split. The result is a training and validation sets with the approximate sizes of $126$ and $54$ thousand samples, respectively. Each sequence comprises $13$ triples of image, beam, and link status. The image-beam pairs in the first 8 triples are used as observed data samples while the link status from the $9$th triple is used as the future ground truth label (i.e., $s_u[t+1]$).
 
\subsection{Evaluation Metrics}
In this section, we present the evaluation metric followed to evaluate the performance of our proposed network architecture. Post training, we evaluate each network on the validation set of ViWi-BT. The primary method of evaluating the model performance is using top-1 accuracy. Th top-1 accuracy is defined as follows:

\begin{equation}
    Acc_{top-1} = \frac{1}{U} \sum_{u=1}^{U} \mathbbm{1} \{\hat s_u[t+1] = s_u[t+1] \},
\end{equation}where $\hat s_u[t+1]$ is the predicted blockage value for user $u$ at time instance $t+1$ when provided with the sequence of observation $S_{u}[t]$ as defined in Section~\ref{sec:prob_form}, $s_u[t+1]$ is the ground-truth value of the same data sample, $U$ is the total number of data samples in the validation dataset, $\mathbbm{1}\{.\}$ is the indicator function, with the value of 1 only when the condition provided is satisfied. 

The objective of predicting the user's blockage status is a binary prediction problem, i.e., either the user is LOS or NLOS in the next time instance. The top-1 accuracy might not reflect the robustness of the trained model. Therefore, we propose using Precision, Recall, and F1-score, along with the top-1 accuracy to evaluate the model's performance correctly.

\subsection{Network Training}
This paper studies the ability of the proposed machine learning network to perform blockage prediction using RGB images and the sequence of previously observed beams. However, to highlight and differentiate the potential of vision-aided link-blockage prediction, we develop a baseline model that performs the same task but without the visual data, using only the beam sequences. The model is simply the recurrent component of the proposed solution described in Section~\ref{subsec:rnn}. It takes in the 8-beam sequences and predicts the link-status.

For the vision-aided experiments with RGB images and beam sequences, we utilize a pre-trained YOLOv3 network as our object detector, and we train the last two components of the proposed solution. In order to have a robust detector, the parameters of YOLOv3 are pre-trained on the COCO dataset \cite{coco}, which is a large-scale image dataset with around 80 object categories. The pre-trained detector is plugged into the proposed architecture. Then, the last two components of the architecture, namely feature embedding and recurrent prediction, are trained jointly using the extended ViWi-BT dataset. A cross-entropy loss function is used in training the proposed architecture as well as the baseline model. All the experiments were performed on a single NVIDIA RTX 2080Ti GPU using the PyTorch deep learning framework.

\begin{figure}[t]
	\centering
	\includegraphics[width=1\linewidth]{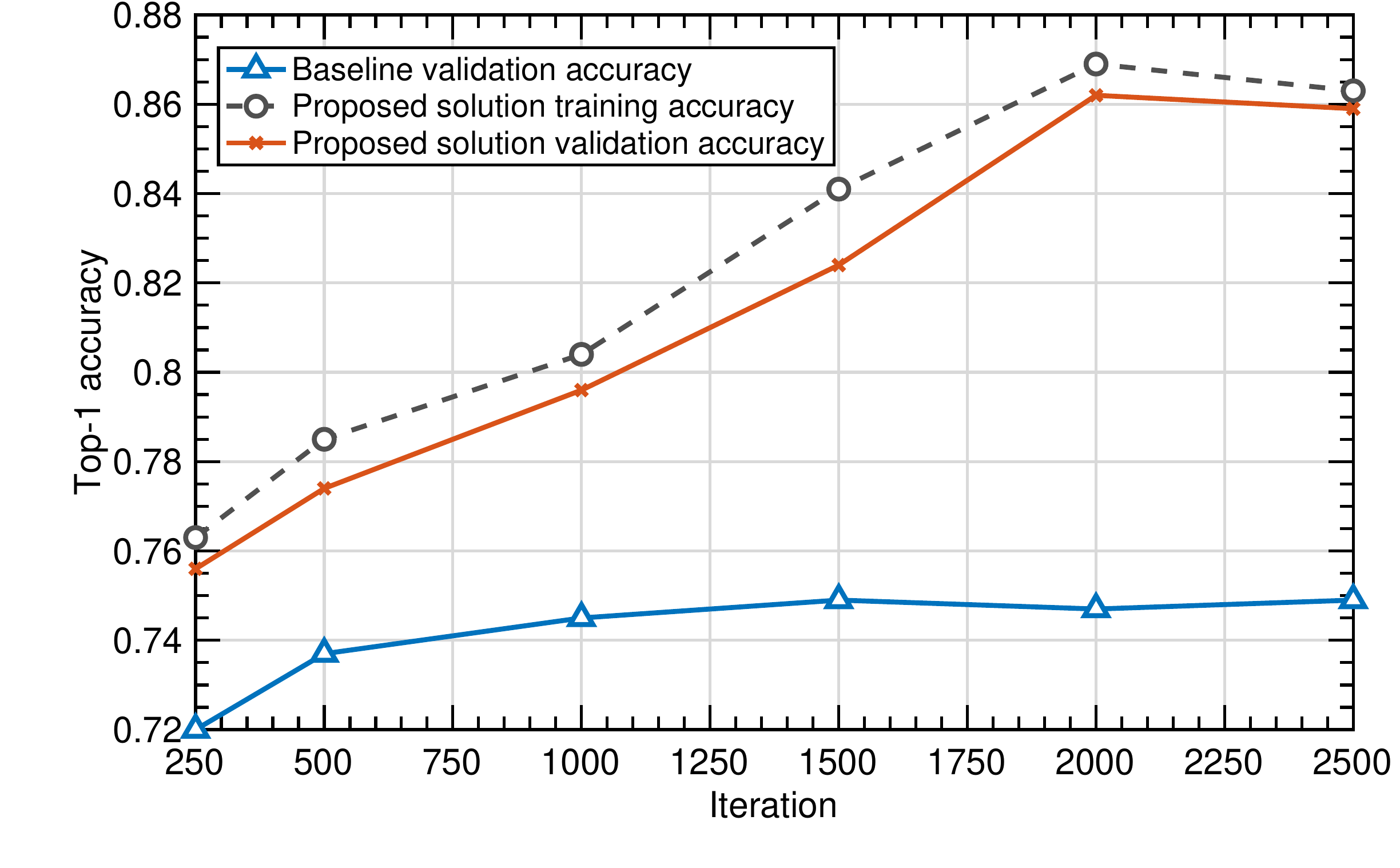}
	\caption{This shows the improvement of top-1 accuracy of the vision-aided future blockage prediction solution as proposed in Section \ref{sec:prop_sol}. The accuracy is reported as a ratio of the correctly predicted sequences to the total number of validation sequences. }
	\label{fig:top-1_acc}
\end{figure}

\begin{table}[]
	\caption{Precision, Recall, F1-score and Top-1 Accuracy Comparison}
	\label{table}
	\centering
	\setlength{\tabcolsep}{5pt}
	\renewcommand{\arraystretch}{1.4}
	\begin{tabular}{|l|c|c|c|c|}
		\hline
		\textbf{Model} & \multicolumn{1}{c|}{\textbf{Precision}} & \multicolumn{1}{c|}{\textbf{Recall}} & \multicolumn{1}{c|}{\textbf{F1-score}} & \textbf{Validation Acc} \\ \hline
		Baseline       & 0.8721                                  & 0.7533                               & 0.808358472                            & 0.7496                  \\ \hline
		Vision-Aided   & 0.8008                                  & 0.9623                               & 0.874153                               & 0.8598                  \\ \hline
	\end{tabular}
	\label{ref:F1-score}
	\vspace{-2mm}
\end{table}

\section{Performance Evaluation} \label{sec:perf_eval}

Identification of the user's existence in the scene and developing an overall understanding of the user's surroundings are integral to an efficient machine learning model for link blockage prediction. Fig.~\ref{fig:top-1_acc} plots the validation top-1 accuracy versus the number of iterations for both the baseline and the proposed solutions. The figure shows that beamforming vectors alone do not reveal enough information about future blockages; the baseline solution achieves around 75\% top-1 accuracy, leaving some room for improvement. The proposed solution, on the other hand, performs very well, providing a $\approx$15\% improvement in top-1 accuracy over that of the baseline solution. Another interesting observation from the figure is that the improvement over the baseline model is observed from the very onset of training. This highlights the importance of visual data in tackling the link-blockage prediction problem. We can also observe from the figure that the proposed solution experience consistent performance throughout training, which is depicted by the small gap between the training and validation accuracies throughout the training process. To further substantiate those results, we present the Precision, Recall, and F1-score in Table.~\ref{ref:F1-score}. The Recall score of $\approx$96\% signifies the model ability in identifying the true cases of blocked links while the $80\%$ precision reflects the confidence of the model in detecting those cases, i.e., how many of the predictions are \textbf{not} falsely identified as blocked links. The F1-score of $\approx$86\% provides a good single-value summary of the precision-recall performance.

\begin{figure}
\centering
\includegraphics[width=\linewidth]{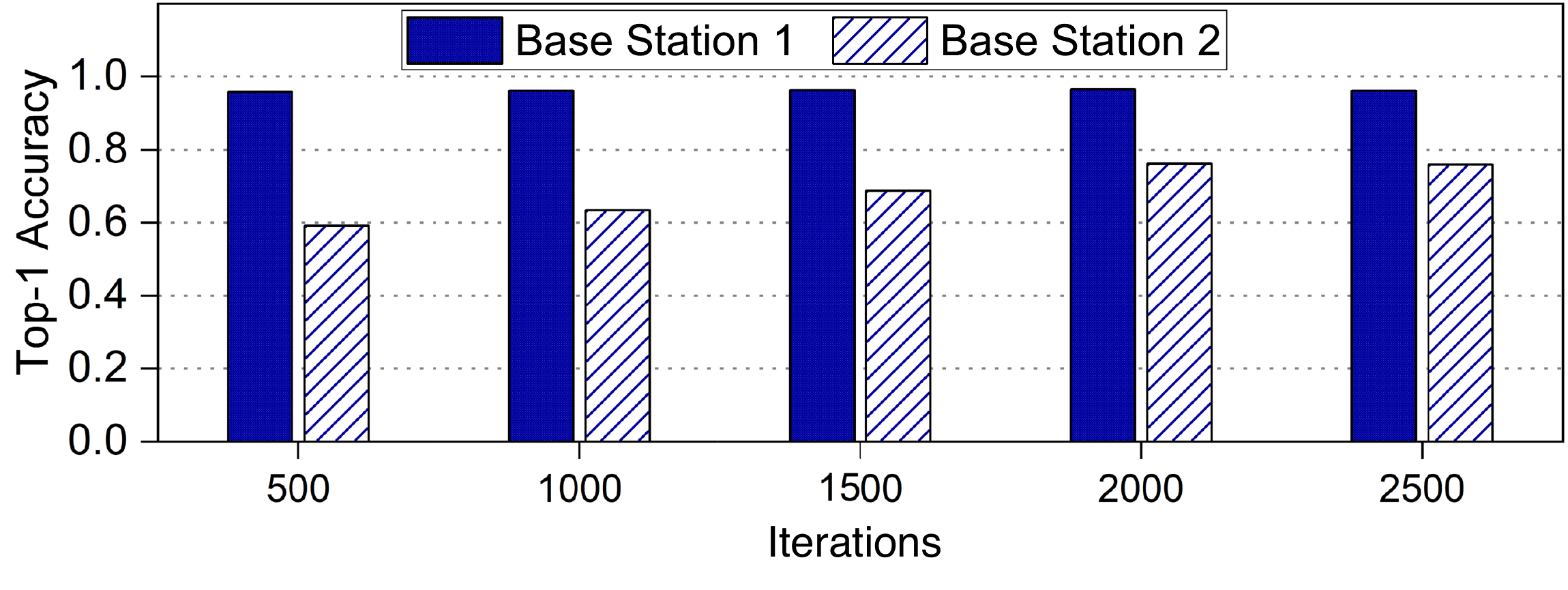}
\caption{The figure presents the individual Top-1 accuracy of Basestation 1 and Basesation 2 as the training progresses.  }
\label{ref:bs_acc}
\vspace{-2mm}
\end{figure}

\begin{figure}
\centering
\includegraphics[width=\linewidth]{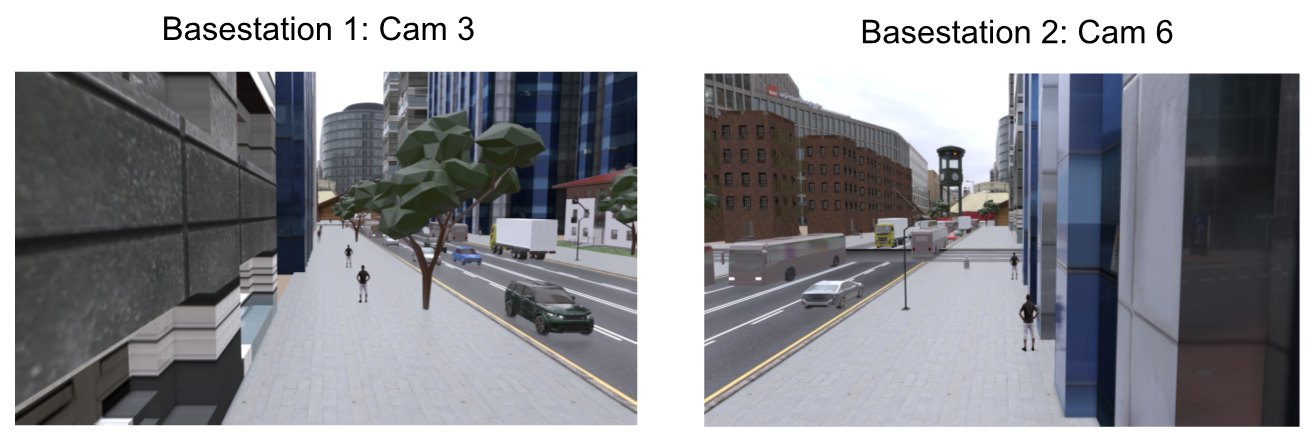}
\caption{This shows the scene captured from Camera 3 of Basestation 1 and Camera 6 of Basestation 2.  }
\label{ref:bs_images}
\vspace{-4mm}
\end{figure}

Fig.~\ref{ref:bs_acc} presents the top-1 validation accuracies versus the number of iterations for the individual basestation in the ViWi dataset. It is observed that basestation 1 performs better consistently as compared to basestation 2. For a machine learning model, it is generally easier to predict the future blockage status if the user is in the central view of a basestation, compared to the mid-peripheral and far-peripheral view of the basestation. As represented in Fig.~\ref{ref:bs_images}, Camera 6 in basestation 2, with a far-peripheral view of the surroundings, presents an uphill task to the machine learning model. However, the performance of that basestation improves consistently as the training iteration increases. 

The overall performance of the proposed model enforces our hypothesis that the additional sensory data like RGB images of the user's surroundings extracted in the form of image embeddings help in identifying possible future blockages, thereby boosting the performance of the model.

\section{Conclusion}
\label{sec:conc}
Enabling low latency and high reliability mmWave and sub-terahertz wireless networks calls for the development of innovative solutions that overcome the various challenges facing those networks. Recent advances in deep learning and computer-vision along with the increasing reliance on LOS connections in mmWave and sub-terahertz networks have motivated the fusion of visual and wireless data to overcome key challenges such as LOS link blockages. In this paper, we proposed a novel deep learning architecture capable of learning proactive link-blockage predictions in a multi-user scenario by leveraging sequences of observed beams and RGB images.
The proposed solution combines the advantages of CNN and RNN networks to develop a sense of scene understanding. Experimental results on the ViWi dataset demonstrate that the proposed approach outperforms the solutions that rely only on beam sequences, which highlights the potential of leveraging visual data in future wireless communication systems.


\end{document}